\documentclass[11pt,a4paper]{article}

\usepackage[T1]{fontenc}
\usepackage[utf8]{inputenc}
\IfFileExists{lmodern.sty}{\usepackage{lmodern}}{\usepackage{mathptmx}}
\usepackage{microtype}
\usepackage[margin=2.5cm]{geometry}
\usepackage{amsmath}
\usepackage{graphicx}
\usepackage{booktabs}
\usepackage{tabularx}
\usepackage{array}
\usepackage{xcolor}
\usepackage{tikz}
\usetikzlibrary{positioning,arrows.meta,fit,backgrounds}
\usepackage{listings}
\usepackage[numbers,sort&compress]{natbib}
\usepackage[hidelinks]{hyperref}
\usepackage[capitalise,noabbrev]{cleveref}
\usepackage{authblk}

\definecolor{ecblue}{HTML}{0072B2}
\definecolor{ecgreen}{HTML}{009E73}
\definecolor{ecorange}{HTML}{E69F00}
\definecolor{termbg}{HTML}{F4F4F4}

\lstdefinestyle{terminal}{
  basicstyle=\ttfamily\footnotesize,
  backgroundcolor=\color{termbg},
  frame=single, rulecolor=\color{black!25},
  columns=fullflexible, keepspaces=true,
  breaklines=true, breakatwhitespace=true,
  xleftmargin=4pt, xrightmargin=4pt, aboveskip=6pt, belowskip=4pt
}

\newcommand{\EC}{\textsc{EasyClassifier}}
\newcommand{\pp}{\,pp}

\title{\EC: Honest, Reproducible Machine-Learning Classification\\ for Researchers Who Do Not Program}

\author[1]{Ahmad B. A. Hassanat\thanks{Corresponding author: \href{mailto:ahmad.hassanat@gmail.com}{ahmad.hassanat@gmail.com}}}
\author[2]{Ghada A. Altarawneh\thanks{\href{mailto:ghadaa@mutah.edu.jo}{ghadaa@mutah.edu.jo}}}
\affil[1]{Faculty of Information Technology, Mutah University, Karak, Jordan}
\affil[2]{Department of Accounting, Mutah University, Karak, Jordan}
\date{}

\begin{document}
\maketitle

\begin{abstract}
Machine-learning classification is now used across medicine, the social sciences, economics, education and engineering, very often by researchers who do not program. Two errors recur in such work: preprocessing learned on all rows before cross-validation (data leakage), and reporting the cross-validation score of the classifier that won a comparison (selection bias). Both make published scores optimistic. We present \EC, an open-source Python package that guides a user from a CSV or Excel file to a finished analysis through a sequence of plain-language, multiple-choice questions, with no code. It makes the correct procedure the default rather than an option: every preprocessing step is learned inside each training fold; the selected classifier receives a separate final score from nested cross-validation (up to 2,000 rows) or an untouched 20\% test set; classifiers run with fixed default parameters and fixed random seeds; and every run writes a report with a ready-to-adapt Methods paragraph, the references to cite, and figures prepared for publication. On ten public datasets from seven fields and a random-label control, split in half so that one half served as an untouched external test, the common practice overstated balanced accuracy by 2.9 percentage points on average (up to 9.9 on small data), whereas the score \EC{} reports had a mean signed bias of +0.7 points and a mean absolute error of 2.3 points (one-sided Wilcoxon $p = 3.2\times10^{-4}$, 30 runs). On random labels it reported 48.8\% balanced accuracy against a chance level of 50\%. \EC{} is available under the MIT license from PyPI (\texttt{pip install easyclassifier}), GitHub and Zenodo.
\end{abstract}

\noindent\textbf{Keywords:} machine learning; classification; no-code; data leakage; nested cross-validation; reproducibility; research software; Hassanat distance

\begin{table}[h]
\centering
\small
\caption{Code metadata.}
\label{tab:metadata}
\begin{tabularx}{\linewidth}{@{}lX@{}}
\toprule
Current version & 0.8.1 \\
Code repository & \url{https://github.com/ahmadhassant/easyclassifier} \\
Archived version (DOI) & \href{https://doi.org/10.5281/zenodo.23122902}{10.5281/zenodo.23122902} \\
Package & \url{https://pypi.org/project/easyclassifier/} \\
License & MIT \\
Language and requirements & Python $\geq$ 3.10; pandas, NumPy, scikit-learn, matplotlib, joblib, openpyxl; optional XGBoost, LightGBM, and a LaTeX distribution for the PDF report \\
Platforms & Windows, macOS, Linux (tested automatically on all three, Python 3.10--3.14) \\
Documentation & User guide for non-programmers: \url{https://github.com/ahmadhassant/easyclassifier/blob/main/docs/USER_GUIDE.md} \\
Support & GitHub issues; \href{mailto:ahmad.hassanat@gmail.com}{ahmad.hassanat@gmail.com} \\
\bottomrule
\end{tabularx}
\end{table}

\section{Motivation and significance}
\label{sec:motivation}

Supervised classification has become a standard tool well beyond computer science. A clinician predicts a diagnosis from routine measurements, an economist predicts loan default, a political scientist predicts turnout, an educator predicts which students will drop out. Many of these researchers know their data and their field thoroughly but do not write code, and few have formal training in how machine-learning results should be evaluated.

Evaluation is where most errors occur. In a survey of the literature, \citet{kapoor2023leakage} found errors from data leakage in 329 published papers across seventeen scientific fields, in some cases leading to strongly over-optimistic conclusions. Two mistakes are especially common and easy to make in good faith:
\begin{enumerate}
  \item \emph{Leakage through preprocessing.} Missing values are filled, columns are scaled or categories encoded using all rows, and the data are only then split for cross-validation. The test folds have thereby influenced the training \citep{kapoor2023leakage,lones2021pitfalls}.
  \item \emph{Selection bias.} Several classifiers are compared by cross-validation and the winner's score is reported as its expected performance. The winner partly won by luck on those particular splits, so its score is optimistic; the remedy is to score the whole selection procedure on data it never saw, by nested cross-validation or an untouched test set \citep{varma2006bias,cawley2010overfitting}.
\end{enumerate}

Graphical environments such as WEKA \citep{hall2009weka}, Orange \citep{demsar2013orange} and KNIME \citep{berthold2009knime} remove the need to program and are rightly popular in teaching and applied work. They are, however, general workbenches: a correct evaluation is possible, but the user has to know how to build it, and the most direct way to compare classifiers in them is to compare their cross-validation scores. Automated machine-learning systems such as auto-sklearn \citep{feurer2015autosklearn} and TPOT \citep{olson2016tpot} search large spaces of pipelines and hyperparameters; they are programming libraries, and the search itself adds a further layer of selection that must be accounted for in the reported score.

\EC{} addresses a narrower question: \emph{how can a researcher who does not program obtain a classification result that is correct, reproducible and ready to report?} Its answer is a guided interview, one question at a time, in which the honest procedure is the default and the user is told in plain words what each result means and which number to report. This paper describes the design (\cref{sec:design}), the software (\cref{sec:software}), a complete example (\cref{sec:example}), an evaluation of whether the reported scores are honest (\cref{sec:evaluation}), and the expected impact and limitations (\cref{sec:impact}).

\section{Design principles}
\label{sec:design}

Five principles guided every decision.

\paragraph{P1 -- No code, no jargon unless wanted.} The user never writes or edits code. Every question offers numbered choices with a recommended default (ENTER accepts it), and any option can be explained on demand by typing \texttt{?N}. Columns are described in words (e.g.\ \texttt{Species -- 3 categories (setosa, versicolor, virginica)}). Three modes, Beginner, Advanced and Research, show progressively more control.

\paragraph{P2 -- Honest by default.} Every operation that learns from data (imputation, encoding, scaling) is fitted on training rows only, inside each validation split. The selected classifier is given a separate final score, and the comparison table is labelled as ``used only for choosing; slightly optimistic, do not report''.

\paragraph{P3 -- Default parameters, no tuning.} Classifiers use the default parameters chosen by their developers. Tuning on one small dataset adds another layer of selection and transfers poorly, while defaults perform reasonably on average for many algorithms \citep{probst2019tunability}. Defaults also make the analysis short to describe and easy to repeat.

\paragraph{P4 -- Exactly reproducible.} All randomness has fixed seeds, so the same data and version always give the same numbers. Every run writes a log of every choice and its own dated results folder.

\paragraph{P5 -- Ready to report.} The output is a report that states what was done and what to report, with a Methods paragraph the user can adapt, the citations needed, and figures prepared to publication standards.

\section{Software description}
\label{sec:software}

\subsection{Architecture}

\EC{} is a pure-Python package built on scikit-learn \citep{pedregosa2011sklearn}, pandas \citep{mckinney2010pandas}, NumPy \citep{harris2020numpy} and matplotlib \citep{hunter2007matplotlib}. It runs in a terminal window on Windows, macOS and Linux and is started with \texttt{easyclassifier} or \texttt{python -m easyclassifier}. The code is organised in independent modules: data loading and inspection, target checks, preprocessing, a classifier registry, evaluation, figures, permutation importance, and reporting (CSV, Excel and \LaTeX/PDF). The wizard calls these modules but contains no modelling logic, so new classifiers, metrics or figures can be added without changing the interface. A test suite of 54 tests runs automatically on every change on the three operating systems and Python 3.10--3.14, once with the oldest supported library versions, and once compiling the PDF report.

\subsection{Workflow}

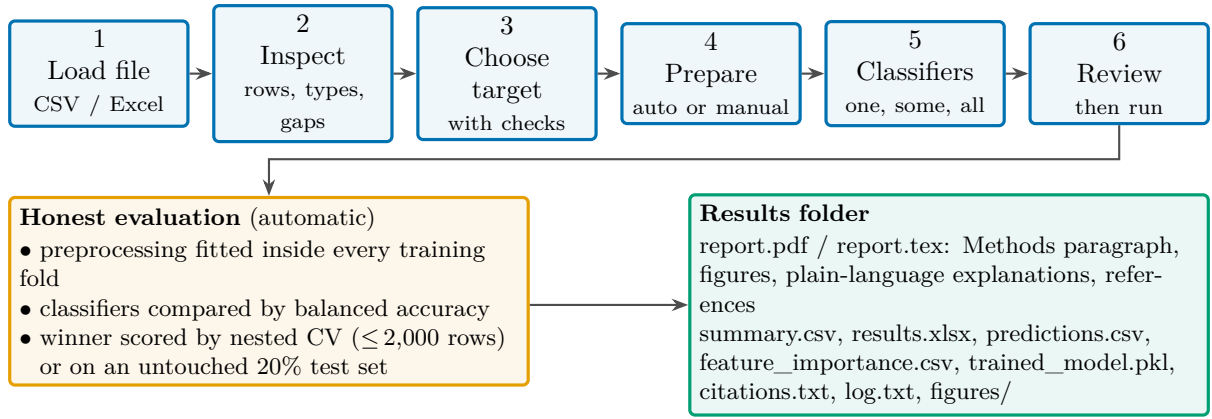
\begin{figure}[t]
\centering
\begin{tikzpicture}[
  node distance=4mm and 3mm,
  stepbox/.style={draw=ecblue, very thick, rounded corners=2pt, fill=ecblue!6,
               align=center, font=\small, minimum height=12mm, text width=21mm},
  outbox/.style={draw=ecgreen, very thick, rounded corners=2pt, fill=ecgreen!8,
              align=left, font=\footnotesize, text width=66mm},
  arr/.style={-{Stealth[length=2.2mm]}, thick, black!70}]
\node[stepbox] (s1) {1\\ Load file\\ \scriptsize CSV / Excel};
\node[stepbox, right=of s1] (s2) {2\\ Inspect\\ \scriptsize rows, types, gaps};
\node[stepbox, right=of s2] (s3) {3\\ Choose target\\ \scriptsize with checks};
\node[stepbox, right=of s3] (s4) {4\\ Prepare\\ \scriptsize auto or manual};
\node[stepbox, right=of s4] (s5) {5\\ Classifiers\\ \scriptsize one, some, all};
\node[stepbox, right=of s5] (s6) {6\\ Review\\ \scriptsize then run};
\foreach \a/\b in {s1/s2,s2/s3,s3/s4,s4/s5,s5/s6} \draw[arr] (\a) -- (\b);

\node[draw=ecorange, very thick, rounded corners=2pt, fill=ecorange!8, align=left,
      font=\footnotesize, text width=66mm, anchor=north west] (eval)
      at ([yshift=-9mm]s1.south west)
  {\textbf{Honest evaluation} (automatic)\\[1pt]
   $\bullet$ preprocessing fitted inside every training fold\\
   $\bullet$ classifiers compared by balanced accuracy\\
   $\bullet$ winner scored by nested CV ($\leq$\,2{,}000 rows)\\
   \phantom{$\bullet$} or on an untouched 20\% test set};
\node[outbox, anchor=north east] (outp) at ([yshift=-9mm]s6.south east)
  {\textbf{Results folder}\\[1pt]
   report.pdf / report.tex: Methods paragraph,\\ figures, plain-language explanations, references\\
   summary.csv, results.xlsx, predictions.csv,\\ feature\_importance.csv, trained\_model.pkl,\\ citations.txt, log.txt, figures/};
\draw[arr] (s6.south) -- ++(0,-4.5mm) -| (eval.north);
\draw[arr] (eval.east |- outp.west) -- (outp.west);
\end{tikzpicture}
\caption{A run of \EC. Each step is one or a few multiple-choice questions with a recommended default. The evaluation and the outputs need no decisions from the user.}
\label{fig:workflow}
\end{figure}

\Cref{fig:workflow} shows the six steps of a run.

\paragraph{Loading and inspection.} The user types a file name or drags the file into the window; \texttt{.csv} (including semicolon-separated files with decimal commas) and \texttt{.xlsx} are read. A built-in demonstration dataset is loaded by typing \texttt{demo}. The inspection reports rows, columns, numeric and text columns, missing values and duplicate rows. Identical rows are flagged as probable copies unless many are expected by chance (few possible value combinations, as in survey answers), in which case they are kept.

\paragraph{Choosing what to predict.} Every column is listed with a plain-language description and the most likely class column is suggested. The choice is checked: a column with a single value is refused; a column that differs in every row (an identifier) triggers a warning; a numeric measurement can be split into two to four equal-sized groups or at a chosen value, and the cut-points are recorded; classes with a single row, which can never be tested, must be removed or another column chosen. Predictor columns that cannot help, such as identifiers and constant columns, are left out with a note.

\paragraph{Preparation.} In automatic mode, rows without a class value and accidental duplicates are removed, and imputation (median, or the most frequent value for text columns), one-hot or ordinal encoding (depending on the number of categories) and scaling are added to the pipeline of each classifier that needs them. Scaling is chosen per classifier: standardisation for SVM, logistic regression and the neural network, scaling to 0--1 for KNN with the Hassanat distance, and none for tree-based methods and naive Bayes. Only steps that learn nothing from the data run before splitting; everything else is part of a scikit-learn \texttt{Pipeline} and is therefore refitted on the training rows of every split.

\paragraph{Classifiers.} Nine classifiers are offered (\cref{tab:classifiers}); pressing ENTER selects all of them. When KNN is chosen, the user picks one of six distances; the Hassanat distance \citep{hassanat2014arxiv,hassanat2014distance,hassanat2014distance2,hassanat2014distance3}, which is bounded per dimension and therefore robust to outliers and to differences in scale \citep{abualfeilat2019distance}, is the default, and the report states which form of its formula (non-negative or signed values \citep{hassanat2014distance2}) was applied, with the citation.

\begin{table}[t]
\centering
\small
\caption{Classifiers in \EC{} 0.8.1. All use library defaults and a fixed random seed.}
\label{tab:classifiers}
\begin{tabular}{@{}lll@{}}
\toprule
Classifier & Implementation & Settings that differ from the library default\\
\midrule
Decision tree & scikit-learn & -- \\
Random forest & scikit-learn & -- \\
Support vector machine & scikit-learn (RBF kernel) & probability estimates enabled (for ROC curves)\\
Logistic regression & scikit-learn & up to 1{,}000 iterations\\
K-nearest neighbours & \EC{} distances & $k = 5$; six distances, Hassanat by default\\
Naive Bayes & scikit-learn (Gaussian) & -- \\
Neural network (MLP) & scikit-learn & up to 500 iterations\\
XGBoost (optional) & xgboost & -- \\
LightGBM (optional) & lightgbm & -- \\
\bottomrule
\end{tabular}
\end{table}

\subsection{Evaluation and the final score}

Classifiers are compared by stratified 5-fold cross-validation (when the user reviews each step manually, hold-out, 10-fold and leave-one-out validation are also offered) and ranked by balanced accuracy, which is defined for any number of classes and is not inflated by class imbalance \citep{brodersen2010balanced}. The selected classifier then receives a separate final score by one of two methods, chosen automatically from the data size:
\begin{itemize}
  \item \emph{Nested cross-validation} (up to 2{,}000 rows): within each of five outer folds the complete comparison is repeated on the training part only, its winner is fitted, and that winner predicts the outer test part. The outer predictions give the final score; the report also states which classifier won in each fold.
  \item \emph{Final test set} (larger data): 20\% of rows are set aside before anything else, the comparison runs on the other 80\%, and the winner is scored once on the untouched 20\%.
\end{itemize}
Advanced and Research modes can override the automatic choice.
The user may select accuracy, precision, recall, F1, macro specificity, balanced accuracy, Matthews correlation, Cohen's kappa, ROC AUC and log loss. Macro specificity averages one-versus-rest specificity across classes; for binary data it equals balanced accuracy, rather than negative-class specificity alone. Column importance is measured by permutation importance \citep{breiman2001rf} on rows not used for training. The saved model is the complete pipeline (preparation and classifier) refitted on all rows, so it can be applied directly to new raw data with the same columns.

\subsection{Outputs}

Each run writes a new dated folder. The report (\LaTeX, compiled to PDF when a \LaTeX{} program is installed) contains a summary, a description of the data, a Methods paragraph describing exactly what was done, the results with a one-sentence explanation of every measure, the figures with guidance on how to read them, column importance and references. Five figures are drawn by default: class distribution, classifier comparison with the score of every test fold, confusion matrix with counts and percentages, ROC curves (one per class for three or more classes) and feature importance. Precision--recall curves are added for imbalanced classes and a missing-value map when data are incomplete; Advanced mode adds a correlation matrix, the most important columns by class and a learning curve. Figures are written at 300\,dpi, optionally also as PDF or SVG, in one of four colour themes (colour-blind safe by default, greyscale with patterns, high contrast, soft), and each class keeps its colour in every figure. The report also says in words whether the winner is clearly better than the runner-up and whether more data would be likely to help.

\section{Illustrative example}
\label{sec:example}

\Cref{lst:session} shows an abridged session on the built-in demonstration data (Fisher's iris flowers \citep{fisher1936iris}), in which the user pressed ENTER at every question.

\begin{figure}[t]
\begin{lstlisting}[style=terminal]
Which column do you want to predict?
    1. sepal_length - numbers from 4.3 to 7.9 (35 different)
    ...
    5. Species - 3 categories (setosa, versicolor, virginica)   <- suggested
  (ENTER = 5, the suggested one)
>> [OK] Target: Species
...
  * Validation: 5-Fold Cross Validation (auto-selected).
  * Final score: nested cross-validation (up to 2,000 rows) (auto-selected).
...
Selected classifier: Neural Network (MLP)

Nested cross-validation estimate - these are the numbers to report:
  * Accuracy           95.30%
  * F1                 95.29%
  * ROC AUC            0.9633
  * Support Vector Machine (SVM) chosen in 2/5 folds; Neural Network (MLP)
    chosen in 1/5 folds; Decision Tree chosen in 1/5 folds; Naive Bayes ...

Comparison of all classifiers (balanced accuracy, ranked) - used only to
choose the winner; slightly optimistic, do not report as the final result:
  * Neural Network (MLP)                96.65%
  * Logistic Regression                 96.64%
  ...
\end{lstlisting}
\caption{Abridged terminal session of \EC{} 0.8.1 on the demonstration data, all defaults accepted. The whole run took under a minute.}
\label{lst:session}
\end{figure}

\begin{figure}[t]
\centering
\begin{tabular}{@{}cc@{}}
\includegraphics[width=0.53\linewidth]{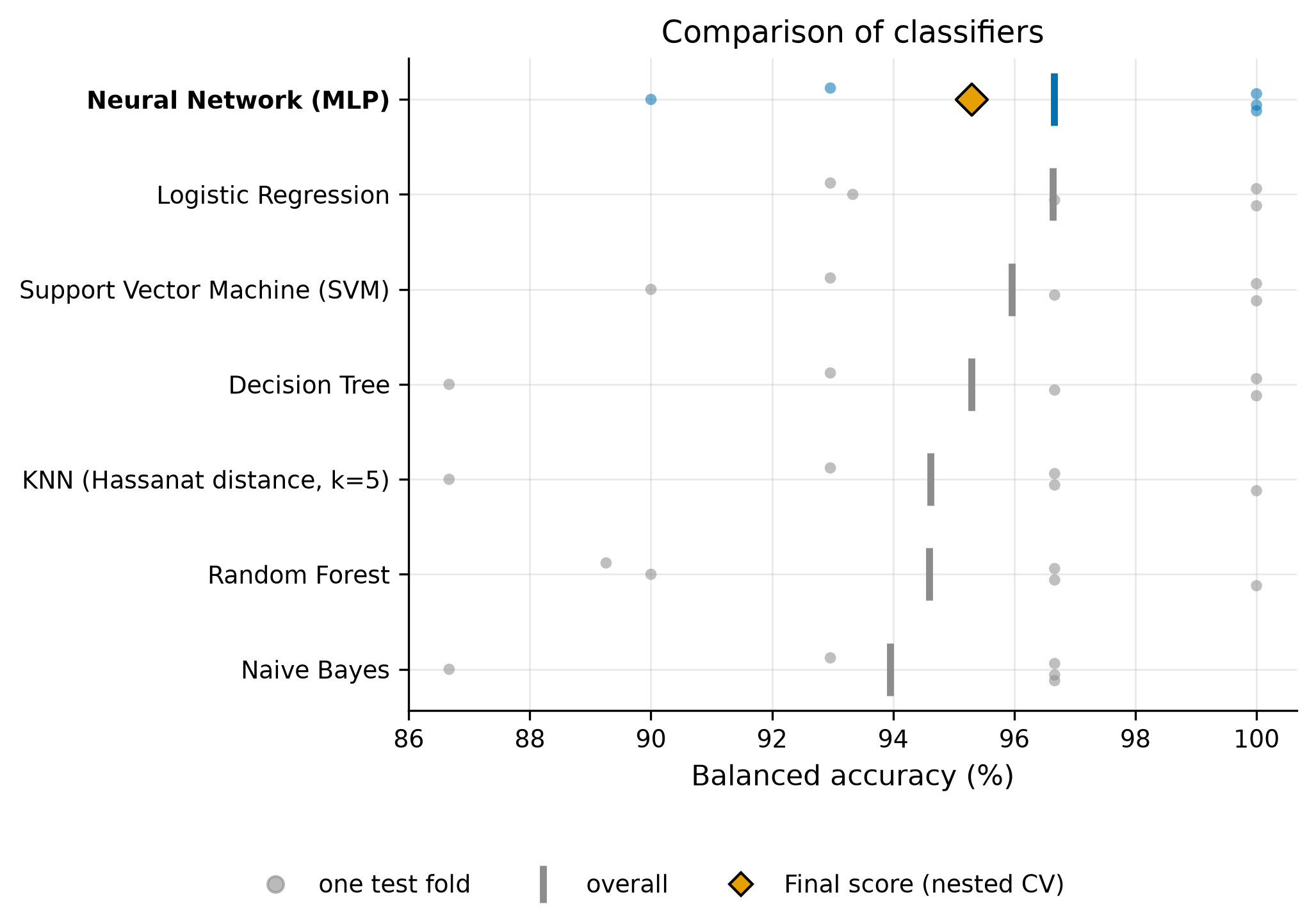} &
\includegraphics[width=0.42\linewidth]{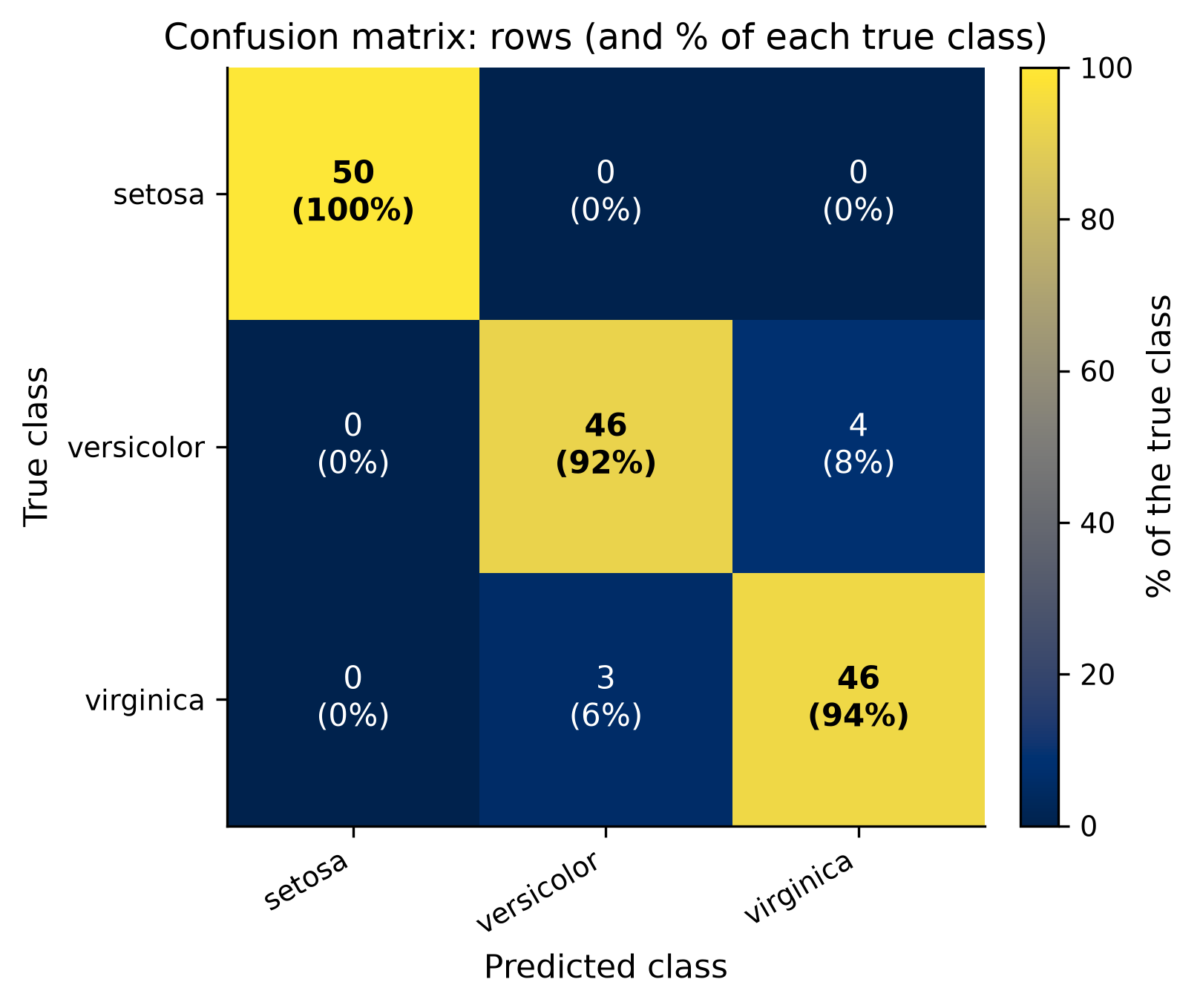}\\
\small (a) & \small (b)
\end{tabular}
\caption{Two of the default figures from the run in \cref{lst:session}, unedited. (a) Classifier comparison: one dot per test fold, a bar for the overall score, and the final nested-CV score of the selected classifier (diamond). (b) Confusion matrix of the nested cross-validation predictions, with counts and the percentage of each true class.}
\label{fig:example}
\end{figure}

Two features of this output matter for non-specialists. First, the number to report (95.3\%) is lower than the winner's comparison score (96.65\%), and the screen says which is which. Second, the outer folds chose four different classifiers, which shows the user that on these data the choice of classifier is largely a matter of chance; the comparison figure (\cref{fig:example}a), which plots the score of every fold, makes the same point visually. The report contains a Methods paragraph that begins:

\begin{quote}\small
Classification was carried out with EasyClassifier version 0.8.1, which is built on scikit-learn. [\ldots] Before modelling, numeric columns were standardised (mean 0, standard deviation 1) for Support Vector Machine (SVM), Logistic Regression and Neural Network (MLP), and scaled to the range 0--1 for KNN (Hassanat distance, k=5). All of these steps were learned from the training data only, separately within every validation split, so that no information from test data entered model training. [\ldots] To avoid optimistic bias from this selection, performance was estimated with nested 5-fold cross-validation, in which the complete comparison was repeated within each outer training fold and the winner was evaluated on the outer test fold.
\end{quote}

\section{Evaluation}
\label{sec:evaluation}

We asked two questions: whether the scores \EC{} tells users to report are honest, and how the KNN distances it offers compare under its automatic preprocessing. The benchmark code, datasets and results are part of the repository and reproduce exactly (\texttt{python benchmarks/run\_benchmarks.py}).

\subsection{Datasets}

We used ten public datasets chosen to represent the fields of \EC's intended users rather than machine-learning benchmarks: medicine (Wisconsin breast cancer \citep{street1993breast}, Pima diabetes \citep{smith1988pima}, low birth weight \citep{hosmer1989logistic}), botany (iris \citep{fisher1936iris}), chemistry (wine cultivars \citep{aeberhard1992wine}), computer vision (8$\times$8 handwritten digits \citep{alpaydin1998digits}), sociology and economics (women's labour-force participation in Canada \citep{fox2011car}), political science (Chile 1988 plebiscite voting intention \citep{fox2011car}; 1992 US election turnout \citep{king2000turnout}) and psychology (volunteering for psychological research \citep{cowles1987volunteers}). They have 149--2{,}523 rows, 3--61 predictors and 2--10 classes; one contains missing values. In the low-birth-weight data the birth weight itself, which defines the class, was removed. An eleventh dataset with 300 rows, ten random predictors and random labels served as a negative control: its true balanced accuracy is 50\% by construction.

\subsection{Benchmark 1: are the reported scores honest?}

\paragraph{Protocol.} Each dataset was split in half, stratified by class, with three different random seeds (33 runs). The first half was analysed exactly as in \EC's automatic mode with all seven standard classifiers; the second half was never touched and gives the true performance of the model each estimate describes, refitted on the whole first half. Three estimates were compared with that truth: (i) the \emph{naive} estimate, the common practice of fitting preprocessing on all rows before cross-validation and reporting the best classifier's cross-validation score (leakage and selection bias); (ii) the best score in \EC's leakage-free comparison (selection bias only); and (iii) \EC's final score. The bias of an estimate is the estimate minus the true performance, in percentage points (pp) of balanced accuracy. All first halves had at most 1{,}261 rows, so every run used nested cross-validation for the final score.

\begin{figure}[t]
\centering
\includegraphics[width=0.82\linewidth]{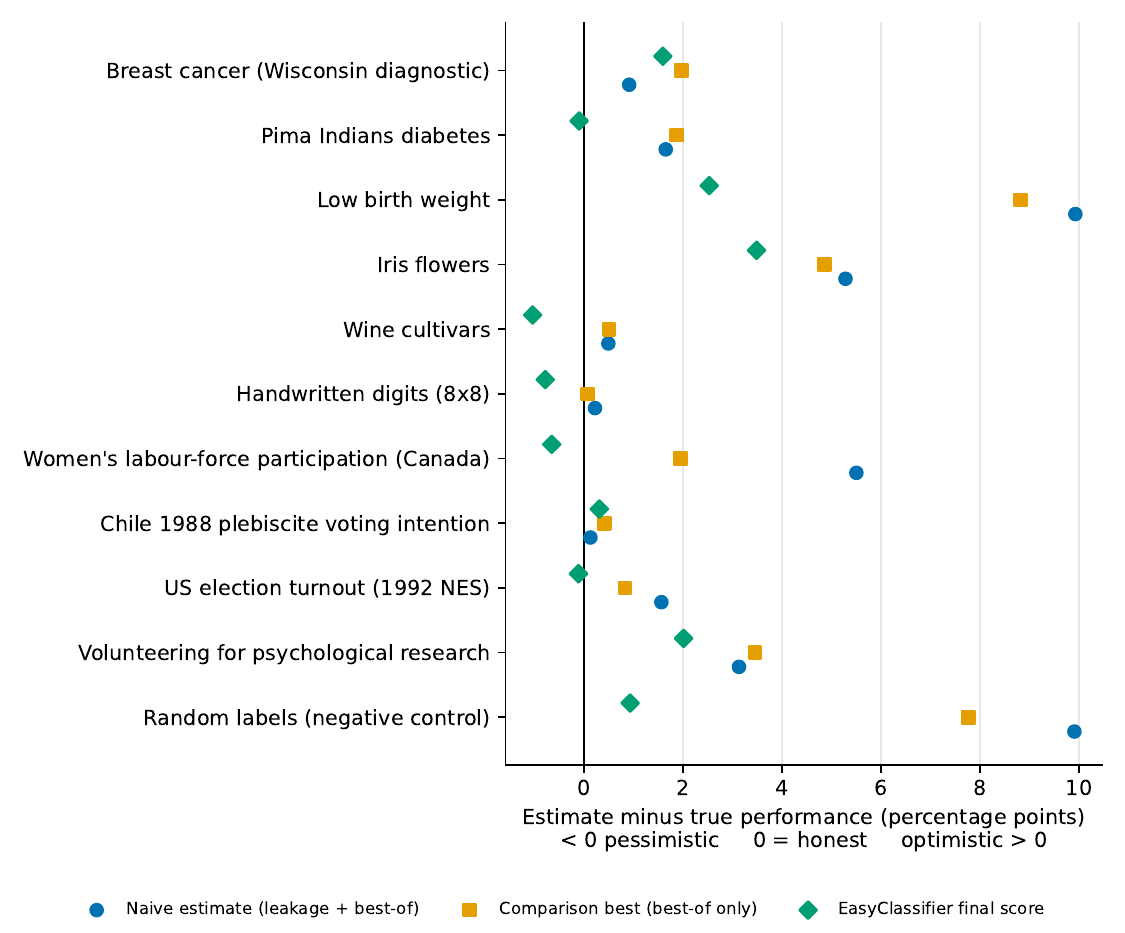}
\caption{Bias of the three estimates (estimate minus performance on the untouched half; mean of three splits). Zero is honest; positive values are optimistic.}
\label{fig:honesty}
\end{figure}

\paragraph{Results.} \Cref{tab:honesty,fig:honesty} show the results. Across the 30 runs on real data the naive estimate was optimistic by 2.9\pp{} on average, the comparison best by 2.5\pp{} and the final score by 0.7\pp{}; the mean absolute errors were 3.0, 2.9 and 2.3\pp. The naive estimate was more optimistic than the final score (one-sided Wilcoxon signed-rank test, $p = 3.2\times 10^{-4}$). Because the three splits reuse each dataset, a sensitivity analysis averaging the discrepancies within each of the ten datasets gave $p = 0.0049$. The bias was largest on small data: on the 183-row low-birth-weight data the naive estimate overstated performance by 9.9\pp, the final score by 2.5\pp. On random labels, where a tool should report chance, the comparison best reported 55.6\% and the naive estimate 52.9\%, while the final score reported 48.8\%. Removing leakage alone (comparison best) reduced the bias only slightly on these data; most of the optimism came from reporting the winner's own score, which is the error users are least likely to know about.

\begin{table}[t]
\centering
\small
\caption{Benchmark 1: balanced accuracy (\%) estimated on the first half of each dataset and measured on the untouched second half (mean of three random splits). The external column is the truth for \EC's selected classifier; the naive estimate is compared with the external score of its own winner when computing its bias.}
\label{tab:honesty}
\begin{tabular}{@{}lrrrr@{}}
\toprule
Dataset & Naive & Comparison best & \EC{} final & External (truth)\\
\midrule
Breast cancer (Wisconsin) & 97.4 & 98.1 & 97.7 & 96.1 \\
Pima diabetes & 73.8 & 73.8 & 71.8 & 71.9 \\
Low birth weight & 63.8 & 63.8 & 57.5 & 55.0 \\
Iris flowers & 97.3 & 96.9 & 95.5 & 92.0 \\
Wine cultivars & 98.0 & 98.4 & 96.8 & 97.9 \\
Handwritten digits (8$\times$8) & 97.5 & 97.3 & 96.5 & 97.3 \\
Women's labour force (Canada) & 53.1 & 51.4 & 48.8 & 49.4 \\
Chile 1988 plebiscite & 51.2 & 51.3 & 51.2 & 50.9 \\
US turnout 1992 & 58.7 & 58.8 & 57.8 & 58.0 \\
Volunteering for research & 54.5 & 54.3 & 52.9 & 50.9 \\
\midrule
Random labels (control) & 52.9 & 55.6 & 48.8 & 47.9 \\
\bottomrule
\end{tabular}
\end{table}

\subsection{Benchmark 2: KNN distances}

\paragraph{Protocol.} KNN with $k = 5$ was evaluated with each of the six distances by repeated stratified 5-fold cross-validation (two repeats) on the ten real datasets, using \EC's automatic preprocessing: columns scaled to 0--1 for the Hassanat distance and standardised for the others. As references we added Euclidean and Hassanat distances on unscaled data and the Hassanat distance on standardised data. Distances were compared by mean rank, the Friedman test and pairwise Wilcoxon signed-rank tests \citep{demsar2006comparisons}.

\begin{figure}[t]
\centering
\includegraphics[width=0.85\linewidth]{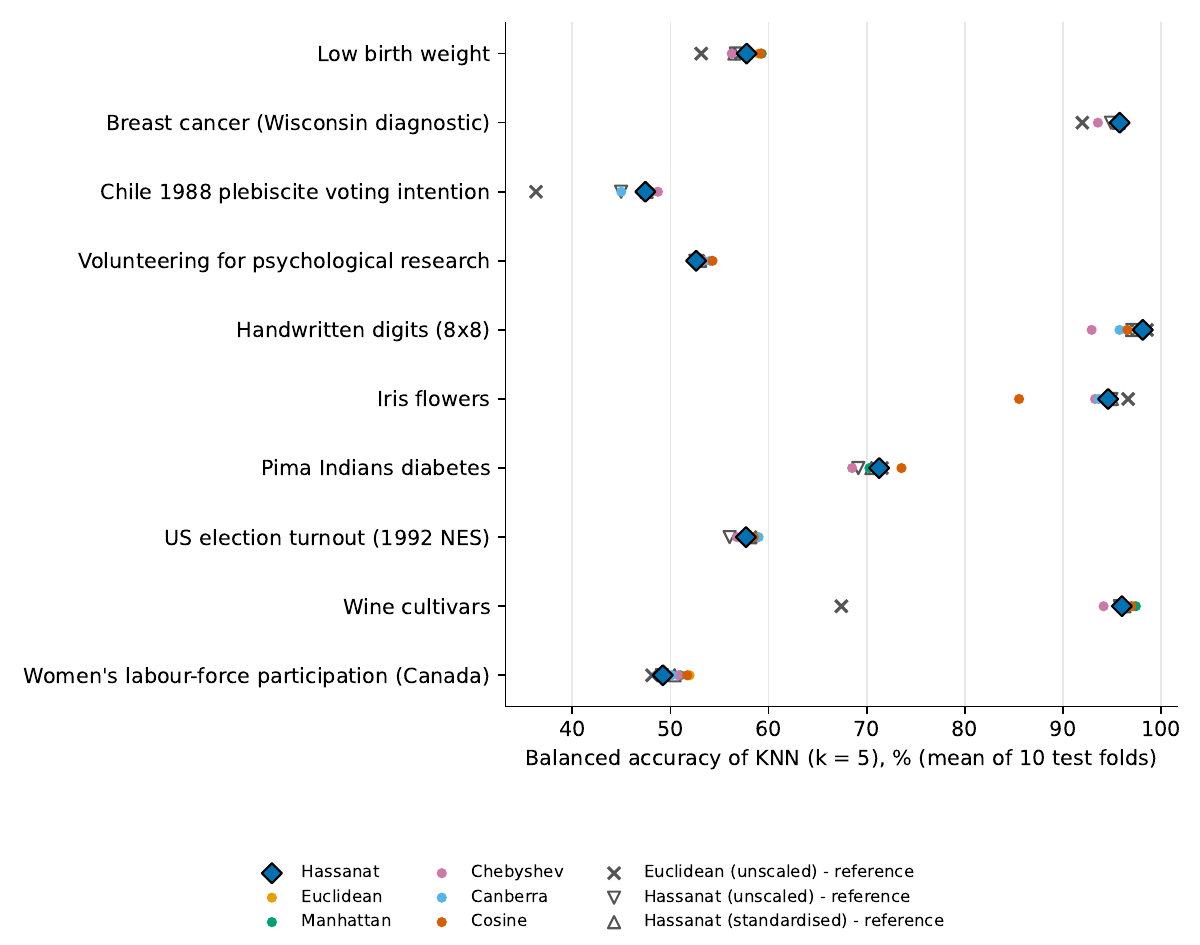}
\caption{Balanced accuracy of KNN ($k = 5$) with each distance under \EC's automatic preprocessing, and three references on unscaled or standardised data (mean of ten test folds).}
\label{fig:knn}
\end{figure}

\paragraph{Results.} With appropriate preprocessing no distance was significantly better than the others (Friedman $p = 0.053$; \cref{fig:knn}). Mean ranks (1 = best) were Euclidean 2.70, Manhattan 2.75, cosine 2.90, Hassanat 3.85, Canberra 3.90 and Chebyshev 4.90. The Hassanat distance was never more than 2.7\pp{} behind the best of the six on any dataset, the smallest worst-case shortfall of the six (Euclidean 2.8, Manhattan 3.3, Canberra 3.7, Chebyshev 5.2, cosine 9.4\pp), although the differences between the first three are small. Scaling to 0--1 improved the Hassanat distance on 9 of 10 datasets compared with unscaled data (mean +0.9\pp, $p = 0.014$), which is why \EC{} 0.8.1 applies it. Without any scaling the picture changes: Euclidean KNN fell to 67.4\% on the wine data and 36.3\% on the Chile data, where the Hassanat distance on the same unscaled columns reached 95.8\% and 45.0\%. This robustness to unscaled and heterogeneous columns, which inexperienced users will often supply, is why the Hassanat distance remains the default; the menu no longer labels it ``recommended'', and users who prefer Euclidean or Manhattan distances lose nothing by choosing them.

\section{Impact, limitations and future work}
\label{sec:impact}

\paragraph{Impact.} \EC{} is aimed at researchers who have data and a question but neither programming experience nor training in machine-learning evaluation: graduate students, clinicians, social scientists, economists, educators and engineers. For them it changes three things. It makes the honest result the easy one, so that the number reported is the number that will hold on new data. It turns the analysis into a short, exactly repeatable procedure that a reviewer can check from the Methods paragraph and the log. And it explains its own output, so that users learn what balanced accuracy, a confusion matrix or nested cross-validation mean while using it. Because the benchmark shows that the largest part of the optimism comes from reporting the winner of a comparison, the tool also serves as a teaching device: on random labels it tells the user, correctly, that nothing can be predicted. For teachers, the built-in demonstration and the plain-language help make it usable in a first course without a programming prerequisite. The package installs with a single command from PyPI; a step-by-step user guide covers installing Python on each operating system.

\paragraph{Limitations.} \EC{} handles classification only; regression, clustering and time series are not supported, and a numeric outcome must be grouped into classes. Hyperparameter tuning and feature selection are deliberately excluded (P3); users who need them should move to a programming environment or AutoML system and account for the additional selection. The interface is a text terminal, which some users find less familiar than a graphical window. Nested cross-validation multiplies the run time by roughly six, which is why it is used only up to 2{,}000 rows; the benchmark above covered that path only, not the 20\% test-set path used for larger data. Column names in non-Latin scripts cannot be typeset by pdf\LaTeX{} and appear as placeholders in the PDF report. Finally, the benchmark used ten modest-sized datasets; it shows the direction and size of the biases but not their exact magnitude in other settings.

\paragraph{Future work.} Planned extensions are regression with the same honest-evaluation design, optional feature selection carried out inside the cross-validation folds, applying a saved model to new data from the wizard, report translation (including Arabic, which requires a Unicode \TeX{} engine), and a graphical front end over the same modules.

\section{Conclusions}

\EC{} lets researchers without programming experience build and evaluate classification models from a spreadsheet, and makes the statistically correct procedure the default: leakage-free preprocessing, a separate final score for the selected classifier, fixed defaults and seeds, and a report that states what to report and how to describe it. On ten public datasets from seven fields its reported scores had a mean signed bias of +0.7 percentage points and a mean absolute error of 2.3 points relative to external performance, whereas the common practice overstated performance by 2.9 points and by up to ten on small data. The software, benchmarks and documentation are openly available, and we welcome reports from users in every field.

\section*{Acknowledgements}
The authors thank the developers of scikit-learn, pandas, NumPy and matplotlib, on which \EC{} is built, and the authors of the public datasets used in the evaluation. Parts of the software, the benchmark scripts and this manuscript were prepared with the help of Claude (Anthropic), an AI assistant, which was used for coding support, drafting and editing. The authors designed the software and the study, checked all code, results and text, and take full responsibility for the content of this paper.

\section*{Software and data availability}
\EC{} 0.8.1 is archived at Zenodo (\href{https://doi.org/10.5281/zenodo.23122902}{doi:10.5281/zenodo.23122902}), distributed on PyPI and developed at \url{https://github.com/ahmadhassant/easyclassifier}. All benchmark datasets are installed with pip; the benchmark code and full results are in the \texttt{benchmarks} folder of the repository. Please cite the software as \citet{hassanat2026easyclassifier}.

\bibliographystyle{plainnat}
\bibliography{references}

\end{document}